%% file: emnlp2023.tex
\pdfoutput=1

\documentclass[11pt]{article}

\usepackage[]{EMNLP2023}

\usepackage{times}
\usepackage{latexsym}
\usepackage{xspace}

\usepackage[T1]{fontenc}

\usepackage[utf8]{inputenc}

\usepackage{microtype}

\usepackage{inconsolata}

\usepackage{amsmath}
\usepackage{graphicx}
\usepackage{subcaption}
\usepackage{booktabs, tabularx}
\newcommand{\framework}{\textsc{Ret-LLM}\xspace}
\newcommand{\nextArg}{\textgreater\textgreater}

\usepackage[most]{tcolorbox}
\tcbset{on line, 
        boxsep=4pt, left=0pt,right=0pt,top=0pt,bottom=0pt,
        colframe=white,colback=gray!15,  
        highlight math style={enhanced}
        }

\newcommand\blfootnote[1]{%
  \begingroup
  \renewcommand\thefootnote{}\footnote{#1}%
  \addtocounter{footnote}{-1}%
  \endgroup
}

\makeatletter
\renewcommand{\@fnsymbol}[1]{\ifcase#1\or \dagger\or \ddagger\or \S\or \P\or \|\else \@ctrerr\fi}
\makeatother
\title{
\framework{}: Towards a General Read-Write Memory for Large Language Models\\
\textcolor{orange}{\it \small Note: This concept paper outlines an initial methodology, now evolved and thoroughly evaluated in \textsc{MemLLM}\xspace.\thanks{\;\;\:MemLLM \citep{modarressi2024memllm}}}
}

\author{
    Ali Modarressi$^{1,2\star}$ ~ Ayyoob Imani$^{1,2\star}$ ~ Mohsen Fayyaz$^{3}$ ~ Hinrich Schütze$^{1,2}$\\
    $^1$Center for Information and Language Processing, LMU Munich, Germany \\
    $^2$Munich Center for Machine Learning, Germany ~
    $^3$Microsoft, Berlin, Germany \\
    \texttt{\{amodaresi, ayyoob\}@cis.lmu.de}
}

\begin{document}
\maketitle

\begin{abstract}

  Large language models (LLMs) have significantly advanced the field of natural language processing (NLP) through their extensive parameters and comprehensive data utilization.
  However, existing LLMs lack a dedicated memory unit, 
  limiting their ability to explicitly store and retrieve knowledge for 
  various tasks. 
  In this paper, we propose \framework{} a novel framework that equips LLMs with a general write-read memory unit, allowing them to extract, store, and recall knowledge from the text as needed for task performance.
  Inspired by Davidsonian semantics theory, we extract and save knowledge in the form of triplets. The memory unit is designed to be 
  scalable, aggregatable, updatable, and interpretable. Through qualitative evaluations, we demonstrate the superiority of our proposed framework over baseline approaches in question answering tasks. Moreover, our framework exhibits robust performance in handling temporal-based question answering tasks, showcasing its ability to effectively manage time-dependent information.
\blfootnote{$^\star$ Equal contribution.}

\end{abstract}

\section{Introduction}

Large language models (LLMs) have significantly advanced the field of natural 
language processing (NLP) in recent years \citep{bubeck2023sparks, chowdhery2022palm, touvron2023llama}.
With their vast parameter count and 
access to extensive data, LLMs have demonstrated remarkable accuracy across various tasks. 
However, current state-of-the-art LLMs lack a dedicated memory unit. 
Instead, they are trained to predict words based on context, encoding knowledge 
implicitly in their parameters, which differs from the ideal memory function.

An ideal memory unit should possess certain characteristics. Firstly, 
it should allow for read and write operations, enabling the language model 
to interact with stored knowledge. Scalability is also crucial, as the memory 
unit should accommodate the consistently evolving nature of knowledge. 
Furthermore, the memory unit should not be limited to textual documents alone; 
it should be capable of acquiring knowledge from diverse sources such as 
database systems. Interpretabilty is desired, granting insight into the 
specific knowledge required by the LLM to solve a given task. Lastly, 
the information stored in the memory unit should be aggregatable, 
enabling the model to combine related information across multiple documents.
For instance an LLM should be able to list all cities of a country mentioned
in multiple documents.

\input{figures/framework_overview.tex}

Previous attempts to incorporate memory into LLMs have fallen short in 
capturing the complete range of memory characteristics.
For example, \citep{zhong-etal-2022-training, wu2022memorizing}
and \citep{chenglanguage}
degrade the memory as the ability to retrieve relevant documents for a 
given query context, and adding them to the context when generating answers.
\citet{park2023generative} merely stores and retrieves previous observations and 
reflections of a generative agent in a simulated environment.

To address these limitations, we introduce \framework{}, (Retentive LLM) 
a solution that endows LLMs with a scalable, updatable, interpretable, 
and aggregatable memory module. Our proposal involves equipping language models
with a memory module, which allows them to extract knowledge from text and 
save it for future reference. When faced with a task, the LLM can query the 
memory module for additional information to support its response. 
The memory module supports updates and can incorporate information 
from non-textual sources such as SQL and no-SQL databases and spreadsheets. Furthermore, 
it enables aggregation of various pieces of information related to a particular concept
scatterred in a huge document or within multiple documents.

Figure \ref{fig:framework_overview} shows the architecture of \framework{}.
It comprises three components: 
an LLM, a controller, and a memory unit. We employ Alpaca \citet{alpaca}, a recently released instruction-tuned language model (LLM), 
and design a fine-tuning process to enable it to acquire the following abilities:
information extraction, information lookup, and fact-based answer 
generation. 

Information extraction entails the identification and extraction of 
triplets in the form of <concept1, relationship, concept2> from 
informative sentences. The information lookup task involves querying the 
memory unit to acquire additional information concerning a given concept 
and its associated relationships when confronted with tasks necessitating 
further information. Lastly, fact-based answer generation involves 
generating a final answer based on the retrieved information.
The triplet-based storage approach draws inspiration from the 
theoretical framework of Davidsonian semantics \citep{davidson1967logical}, 
which provides a foundation for representing concepts described in 
sentences using a triplet-like structure of <event, subject, object>.

The memory module stores the triplets and their vector 
representations. During retrieval, it first searches for an exact 
match of the query text and resorts to a fuzzy search based on 
vector representations if no exact match is found. For efficient 
fuzzy search and retrieval, we employ LSH-based hashing of vector 
representations. The controller acts as an interface, automating 
interactions between users, the LLM, and the memory module, 
ensuring a seamless interaction experience with an intelligent 
chat system.

Our proposed approach offers several advantages over previous methods. 
It enables LLMs to explicitly store and retrieve knowledge, 
which is crucial for real-world NLP applications. 
By incorporating explicit knowledge storage and retrieval, we gain 
better understanding of the workings of these models and the knowledge 
they rely on to solve tasks. The use of an external memory unit separate 
from the LLM ensures scalability and easy modification of stored information. 
The fuzzy search technique enables efficient retrieval of relevant 
information, even in the absence of exact matches. Storing information 
in triplets facilitates the generation of precise and comprehensive solutions, 
particularly when data aggregation is necessary. Lastly, the memory module 
allows for easy incorporation of information from diverse sources and accommodates 
changing facts over time.

Over a qualitative evaluation using question answering examples, we demonstrate cases where a comparable LLM such as Alpaca-7B fails to return a correct answer.
We show that this shortcoming occurs while the model has access to all the information required for generating a valid answer.
However, in our proposed approach after storing the extractable knowledge from the context, the \framework{} shows its capability in answering a question without the need of reinputting the context.
We also demonstrate that \framework{} could handle temporal based QA examples. Since it is equipped with a modifiable memory which could handle temporal facts.




\section{Related Works}

Prior works in the field have explored incorporating relevant context 
into large language models (LLMs) by retrieving and adding relevant 
documents to the task's context. 
\citet{zhong-etal-2022-training} propose training 
LLMs with memory augmentation by introducing trainable memory units 
that are optimized during the training process.
\citet{wu2022memorizing} presents 
the Memorizing Transformer, which can attend to longer documents during 
inference. This approach stores (Key, Value) pairs, extracted from a transformer layer,
in a memory and 
retrieves relevant pairs to add them to the current context during generation. 
\cite{chenglanguage} encode each documents, save them, and retrieve relevant
documents based on 
the current context.
In contrast to these approaches, our method offers improved scalability as 
we do not modify the architecture of the LLM. Instead, we suggest extracting 
and saving information from documents, allowing for the aggregation of extracted 
information from multiple sources. This enables us to provide more relevant and 
concise retrieved information that is closely aligned with the specific question 
being addressed.

\citet{park2023generative} 
utilizes an LLM within a generative agent framework to facilitate the storage 
and dynamic retrieval of a comprehensive record of the agent's experiences 
using natural language. However, there exists a fundamental distinction between 
their architecture and ours. In Park's framework, the memory component is an 
inherent part of the agent itself, while the LLM serves as an external tool 
employed solely for planning the agent's behaviors. Consequently, the LLM 
lacks control over the specific content to be stored and retrieved within 
the agent's memory.

\citet{dhingra-etal-2022-time} contribute to the field by curating a dataset 
specifically designed to differentiate between temporal and non-temporal facts. 
They propose training language models on temporally annotated data to enhance 
their temporal awareness. This work aligns with our research focus on addressing 
temporal information challenges. However, in our proposed solution, we address these 
challenges by introducing an updatable memory module.

\citet{schick2023toolformer} present a methodology that empowers LLMs to 
leverage external tools by generating API calls to access additional 
functionalities, such as using a calculator for task execution. Our 
work shares similarities with their approach in terms of teaching the 
LLM to utilize an external tool. However, it should be noted that our 
focus lies on incorporating a more intricate and influential tool, 
namely the memory module, which has the potential to significantly impact 
the LLM's output.

\section{Approach}
We aim to design a \framework{} where the user can perform two actions: (1): Provide one or a series of informative statements where the \framework{} should
be able to memorize the containing information. Previous methods
perform this task by either training/fine-tuning the LLM over the provided document
or creating a vector representation for the document and storing the representation.
(2): Asking related questions which the \framework{} would answer based on the stored memory.
All these actions should function in a seamless setting where the user should only interact in natural language.

Our \framework{} is constituted by three main components: (1) Controller, (2): Fine-tuned LLM \& (3): Memory.
As shown in Figure \ref{fig:framework_overview}, the controller moderates the flow of information between the user, the LLM and the memory.
The LLM acts as a processing unit, where it receives the texts passed by the controller and figures where it needs to invoke a memory call or not.
Since the LLM operates with text, inspired by \citet{schick2023toolformer}, we standardized the memory calls by implementing a text-based API schema.
Therefore the LLM could generate memory API calls and the controller could apply the LLM API calls to the memory.
In our setting, the memory stores data in triplets by using a three-columned table.
This is based on the theoretical framework of Davidsonian semantics \citep{davidson1967logical}, where concepts described in sentences could be stored in a structure of <first argument, relation, second argument>. 

In the following we describe \framework{} in more detail. The memory-API, how we finetune the LLM to become capable of these calls and the memory structure.
\input{figures/approach.tex}

\subsection{Memory Structure}
Each triplet defines a relationship between two arguments with the following format: $\langle t_1, t_2, t_3 \rangle$ where $t_1$ is the first argument, $t_2$ is the relation and $t_3$ is the second argument in the relationship.
For instance in the sentence: ``Mark Zuckerberg is the CEO of Meta Inc.'' the informative triplet that could be extracted is: $(\text{Mark Zuckerberg},\text{CEO},\text{Meta Inc.})$.

To store these triplets we use a three-columned table where each column is associated with each part of the triplet.
Alongside saving the texts, we store the average representations so that the memory could also handle queries which have semantically similar words.
If the memory module fails to find the exact text in the table, 
it checks for similar texts by comparing the vector representation
of the query text with vector representations of text peices already stored in the dataset.
Therefore for every $t_i$ the mean representation retrieved by the LLM ($\boldsymbol{h}_{AVG}(t_i)$) is stored in a Locality-Sensitive Hashing (LSH) table.
The reason of utilizing LSH is to reduce the computation required for finding similar representations.
Without a hash table for a given query representation, the distances to all of the stored representations should be computed which would be a computationally-expensive task.

\paragraph{Handling Memory Queries.}
\label{sec:handling_mem_queries} 
In a memory query, one or two of the triplet parameters should be provided as input:
\begin{equation*}
  \mathcal{Q} \in \{\langle q_1\rangle, \langle q_2\rangle, \langle q_3\rangle, \langle q_1, q_2\rangle, \langle q_1, q_3\rangle, \langle q_2, q_3\rangle\}
\end{equation*}
Where $q_i$ is the search term for the $i$-th parameter in the stored tuples. 
Before retrieving the query results, each search term is checked 
For a given $\mathcal{Q}$, first the memory checks whether the search terms ($q_i$) have an exact match in the storage table.
If $q_i$ does not exist in the stored terms, we use its average representation $\boldsymbol{h}_{AVG}(q_i)$ and the LSH table for an alternative term ($\tilde{q}_i$) that has an exact match in out memory table.
Possibly, the LSH table may not find an alternative term for the given representation, therefore the query would not have a result: $\mathcal{Q} \rightarrow \emptyset$.
In any case (exact match or similar match), 
the query might have multiple matches in the data table 
($q_i = t_i$).
In this case all resulting triplets would be returned as the query output.

\subsection{Memory-API \& Dataflow}
\label{sec:memory_api}
To enable communication between the memory and the LLM,
we design an API schema for memory read and write functions.
This API allows the controller to understand when the LLM is calling 
the memory and what parameters should be passed.
Based on the triplets discussed in the previous section, the two memory calls are as the following:
\begin{itemize}
  \item \texttt{[MEM\_WRITE\{$t_1$>>$t_2$>>$t_3$\}]}: 
  This structure is for storing a triplet $\langle t_1, t_2, t_3 \rangle$. 
  Depending on the prompt, multiple write calls could be sequentially generated by the LLM to store multiple triplets extracted from a text.
  \item \texttt{[MEM\_READ\{\_>>\_>>\_\}:}$\tcbhighmath{\texttt{\{$t_1$>>$t_2$>>$t_3$\};...]}}$:
  In a memory read, as shown in the API, there are three placeholders that based on $\mathcal{Q}$ atleast one of them should be filled with the search terms.
  Based on the query results from the memory, one or a list of triplets could be returned as shown in the highlighted segment. 
\end{itemize}

Figure \ref{fig:framework_workflow} demonstrates how \framework operates using the memory-API.
Depending on the input given by the user, \framework{} either have to read or write information from or to the memory.
If the user prompt an informative statement (or ideally a full document), it would be memory write scenario.
On the other hand, by having a question in the input, we consider this to be a memory read case.
In both cases the user input is the first input to \framework that is passed on to the LLM.

Based on the given input the LLM infers and generates the relevant API call.
With a memory write case, after the API call is generated the controller detects it and invoke a memory storation function with the given parameters.
The memory receives the data in a triplet format and stores it for future usage.
If a memory read call is generated by the LLM, the controller also detects it and pauses the model's sequence generation for the memory retrieval.
It uses the parameters given inside the read call as the query terms and passes them to the memory.
The memory lists all stored triplets that feature the given search terms (or a semantically similar version of them according to \S \ref{sec:handling_mem_queries}) and return the results back to the controller.
Using the API discussed in the beginning of this section, the read results are listed after the call so that the LLM could use them to produce a naturally sounded answer.
After the answer is produced it is returned back to the user.

As the controller is in between of the user and the LLM, it could hide the whole memory-API schema. 
This would make the user feel an end-to-end simple language modeling experience without knowing the memory functionality behind the scene.

\subsection{Finetuning the LLM}
\label{sec:dataset_and_finetuning}
In this part we discuss how the LLM is finetuned to be capable of generating memory-API calls.
In the end the LLM should be capable of detecting which type of memory call (read or write) it should provoke based on the input.
As stated in Section \ref{sec:memory_api}, the LLM's input may have one of the two previously discussed structures depending on the memory function.
Therefore the LLM should be able to generate and handle this API to store or read the relevant information.
To this end, we develop a synthetic dataset to train the LLM.
The synthetic task is to learn the relationships of the discussed people with the respective corporations.
Based on the stored information, \framework{} should be capable of answering any questions regarding the people, the corporations or the relationships.

\input{tables/mem_read_example.tex}
\input{tables/mem_write_example.tex}
We use a set of firstname and lastnames to generate a synthetic population, called $\mathcal{P}$.
Each person from this population $per \in \mathcal{P}$ could have only one relationship from the following list: $rel \in \mathcal{R}=\{\text{employment}, \text{manager}, \text{investor},$
$ \text{founder}, \text{customer}\}$ with an organization $org \in \mathcal{O}$.
Where $\mathcal{O}$ is a set of corporation names.
Hence, each triplet would be as: $(per, rel, org)$. For instance: $\langle\text{Dominick Alphonso}, \text{employment},$
$ \text{BMW}\rangle$.\footnote{Even though the corporation names are real, the people names are entirely random generated. No identification with actual persons is intended or should be inferred.}
Based on this triplet we can build three \emph{triplet-specific} questions:
\begin{itemize}
  \item $\mathcal{Q}=\langle per \rangle$, e.g. ``Who is Dominick Alphonso?''
  \item $\mathcal{Q}=\langle per, org \rangle$, e.g. ``How Dominick Alphonso is related to BMW?''
  \item $\mathcal{Q}=\langle per, rel \rangle$, e.g. ``Dominick Alphonso is employed by which company?''
\end{itemize}
and the answer to all above should be ``Dominick Alphonso is employed by BMW.''. 
Alongside these questions three other types of questions could be asked that could be relevant to multiple triplets:
\begin{itemize}
  \item $\mathcal{Q}=\langle rel \rangle$, e.g. ``Who are the employees?''
  \item $\mathcal{Q}=\langle org \rangle$, e.g. ``Who are related to BMW?''
  \item $\mathcal{Q}=\langle rel, org \rangle$, e.g. ``Who are employed by BMW?''
\end{itemize}
Unlike the first three, each of these questions could have multiple persons related to the answer.
For each of these questions we expect the model answer the questions without any extra information (e.g. stating the corporation of employment when its not asked).
To create a training data instance from these questions based on the memory-API, we use the templates stated in Table \ref{tab:mem_read_data_template}.
During finetuning the Question, API query (with the \texttt{MEM\_READ} command), API Response and the answer are concatenated as the data input for the LLM.
However, the langauge modeling loss is only applied to the API query and Answer sections.
Since these two segments are the text sequences that the LLM is expected to generate based on the other two segments (Question \& API Response) that are provided by the controller.

As we also need informative examples where have \texttt{MEM\_WRITE} calls, we use a similar strategy by using the population, organizations and relations that were previously defined ($\mathcal{P}, \mathcal{Q}, \mathcal{R}$).
Based on the memory-API, in a memory write scenario the \framework receives a sentence which here contains a relationship information and then the LLM should generate the corresponding memory write calls.
In our dataset we opted to build examples where it states about multiple people whom have the same relationship with the same company: $(per_i,rel,org)$.
The template for the memory write data examples are shown in Table \ref{tab:mem_write_data_template}.
Similar to the question-based examples, the statement and the API call are concatenated to form the full input sequence.
Also the loss function is applied only to the API segment, since the first part is provided by the controller.

We opted to use the Instruction-following Alpaca-7B model \cite{alpaca} as a base model for our finetuning. 
To execute the training in a resource limited setup, we use low-rank adaptation (LoRA) \cite{hu2022lora}. \footnote{The code for finetuning a llama-based model using LoRA is available at: \href{https://github.com/tloen/alpaca-lora}{github.com/tloen/alpaca-lora}}
This parameter efficient measure allows us to finetune the base model on a single A6000 48GB GPU.

\input{figures/eval_example_1.tex}
\input{figures/qa_example.tex}

\section{Qualitative Results}
In this part, we present the internal process and final output on multiple evaluation examples.
These examples were generated with the same procedure stated in \S\ref{sec:dataset_and_finetuning}.
First to demonstrate the importance of our approach, we provide the same example to our base model (Alpaca-7B) in a zero-shot setting.
The input would be a short instruction for the task, the informative sentences from the example and in the end is the question.
As shown in Figure \ref{fig:eval_example_1}, the zero-shot result from the instruction tuned model is clearly incorrect.
While the model does have all the information in its context, its still produces an incorrect response.

In thie same example, the \framework first stores the extracted triplets from the examples into the memory.
After storing the extracted relationships, the \framework could respond to the same question even without having the information in the input.
With the help of the memory-API and the memory itself, the relevant triplet is found. The LLM manages to answer correctly after appending the query result to the memory call.

One potential use cases of our approach is in answering questions that have a temporal context.
For example, the presidency of the United States undergoes a change every 4 to 8 years.
A normal PLM model answers the question about the presidency based on its own training data.
While model retraining or parameter editing has its own challenges, our approach could provide an easy and interpretable solution for this issue (Figure \ref{fig:qa_president_example}).

\section{Conclusion \& Future Work}
In this work, we introduced a \framework capable of storing information and retrieving it in further use.
With a triplet based memory structure, information are stored in relationships between two arguments with a known relation.
The memory could be utilized via a memory-API which is generated by a finetuned LLM.
Using a controller, all components could communicate with each other and the user would interact with the controller being unbeknown of the behind process.
We have shown that the LLM generates the proper API calls in some question answering examples without having the information in its input context.
As this work is still under development, in our next revision we will add a more in-detail empirical evaluation, preferrably on a real dataset.
We also seek to improve our finetuning method to a more generalized setting so that it could be capable of working with more types of informative relations.

\bibliography{anthology,custom}
\bibliographystyle{acl_natbib}

\appendix

\section{Extra Evaluation Example}
\label{sec:extra_eval_example}
\input{figures/eval_example_2.tex}

\end{document}

%% file: figures/framework_overview.tex
\begin{figure}[t]
\centering
    \includegraphics[width=\linewidth, trim=0 0 0 0, clip] {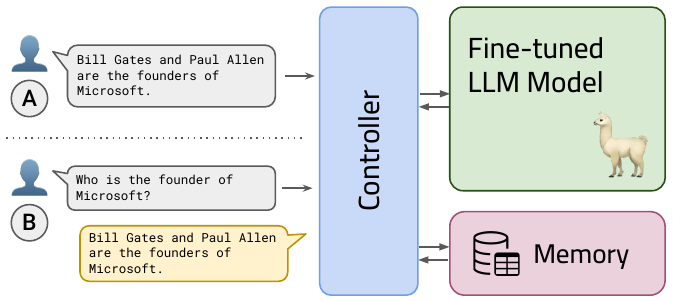}
    \caption{
        An overview of \framework{}. A user could prompt with
        (A): an informative sentence and our approach stores potent information from it inside the memory or
        (B): a question where previously saved information should be utilized to generate a valid answer. 
    }
    \label{fig:framework_overview}
\end{figure}

%% file: figures/approach.tex
\begin{figure*}[t]
    \centering
    \begin{subfigure}[b]{0.97\textwidth}
        \includegraphics[width=\linewidth, trim=2 0 -1 0, clip] {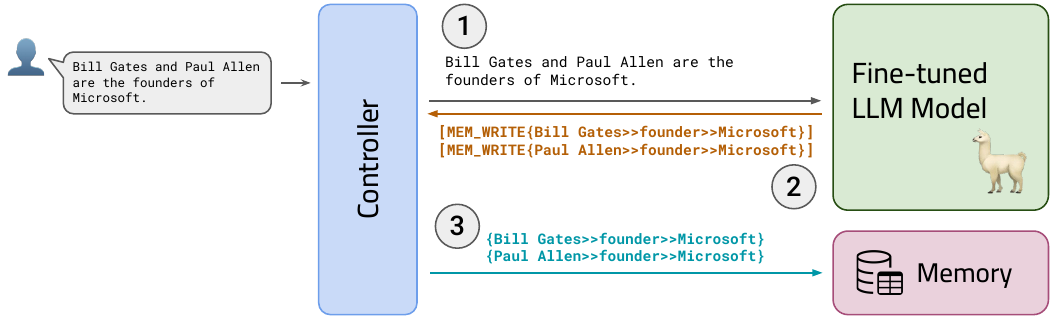}
        \caption{Memory-Write scenario: (1) Controller passes the input to the LLM (2) which generates the appropiate memory write call. (3) The controller gives the data (and their average represntations) to the memory to be stored.}
     \end{subfigure}
     \begin{subfigure}[b]{0.97\textwidth}
        \includegraphics[width=\linewidth, trim=0 0 0 0, clip] {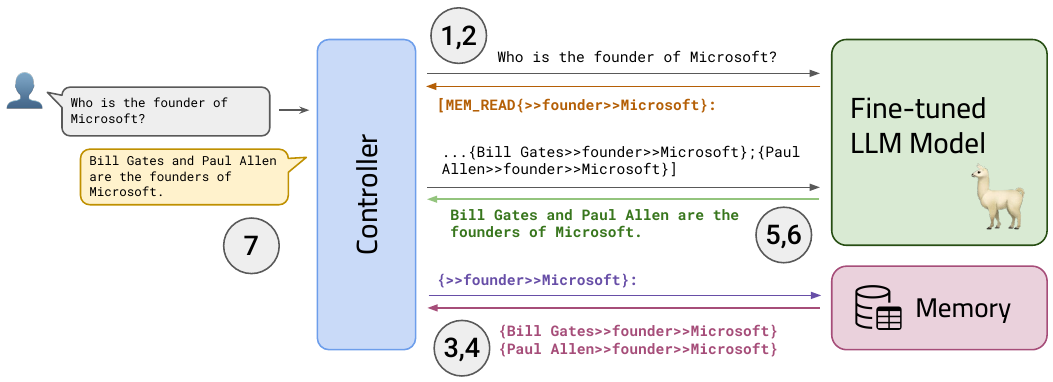}
        \caption{Memory-Read scenario: (1) Controller passes the question to the LLM (2) which generates the appropiate memory read call. (3) The controller apply the query on the memory with the given search terms from the LLM. (4) The memory returns the query results which are (5) forwarded back to the LLM. (6) The LLM generates the answer to the question using the query results and (7) the answer would be returned back to the user.}
     \end{subfigure}
    \caption{A visualization of the process in both read- and write-based inputs.}
    \label{fig:framework_workflow}
\end{figure*}

%% file: tables/mem_read_example.tex
\begin{table*}[]
\begin{center}
    \resizebox{0.97\textwidth}{!}{
    \setlength{\tabcolsep}{9pt}
    \begin{tabular}{l|l|l|l|l}
        \toprule
        \textbf{Query Type}                 & \textbf{Question}                                     & \textbf{API Query}         & \textbf{API Response}          & \textbf{Answer}                                              \\
        \midrule
        $\langle per \rangle$      & Who is \texttt{per}?                         & \texttt{\{per\nextArg\nextArg\}:}    & \texttt{\{per\nextArg rel\nextArg org\}}   & \texttt{per} is \texttt{rel} to \texttt{org}.       \\
        $\langle per, org \rangle$ & How \texttt{per} is related to \texttt{org}? & \texttt{\{per\nextArg\nextArg org\}:} & \texttt{\{per\nextArg rel\nextArg org\}}   & \texttt{per} is \texttt{rel} to \texttt{org}.       \\
        $\langle per, rel \rangle$ & \texttt{per} is \texttt{rel} which company?  & \texttt{\{per\nextArg rel\nextArg\}:} & \texttt{\{per\nextArg rel\nextArg org\}}   & \texttt{per} is \texttt{rel} to \texttt{org}.       \\
        $\langle org \rangle$      & Who are related to \texttt{org}?             & \texttt{\{\nextArg\nextArg org\}:}    & \texttt{\{per$_1$\nextArg rel$_1$\nextArg org\};\{per$_2$\nextArg rel$_2$\nextArg org\};...} & [\texttt{per}$_1$, \texttt{per}$_2$, ...] is\textbackslash are related to \texttt{org}.      \\
        $\langle rel \rangle$      & Who are the \texttt{rel}?                    & \texttt{\{\nextArg rel\nextArg\}:}    & \texttt{\{per$_1$\nextArg rel\nextArg org$_1$\};\{per$_2$\nextArg rel\nextArg org$_2$\};...} & [\texttt{per}$_1$, \texttt{per}$_2$, ...] is\textbackslash are \texttt{rel}.                 \\
        $\langle org, rel \rangle$ & Who are \texttt{rel} \texttt{org}?           & \texttt{\{\nextArg rel\nextArg org\}:} & \texttt{\{per$_1$\nextArg rel\nextArg org\};\{per$_2$\nextArg rel\nextArg org\};...} & [\texttt{per}$_1$, \texttt{per}$_2$, ...] is\textbackslash are \texttt{rel} to \texttt{org}. \\
        \bottomrule
    \end{tabular}}
\end{center}
\caption[]{Memory read data examples for finetuning. 
The first three types of questions are based on a single triplet therefore the API response would be only one triplet.
However the second three may have multiple relevant tiplets stored in the memory as shown in their API-Resonse. Thus, the answer should combine the triplets data into a single sentence.
[\texttt{per}$_1$, \texttt{per}$_2$, ...] is the placeholder of the names written sequentially in a natural way.
For instance: ``Dirk Alosa, Ty Baumkirchner, and Vera Bayless''}
\label{tab:mem_read_data_template}
\end{table*}

%% file: tables/mem_write_example.tex
\begin{table*}[]
\begin{center}
    \resizebox{0.97\textwidth}{!}{
    \setlength{\tabcolsep}{9pt}
    \begin{tabular}{l|l|l}
        \toprule
        \textbf{Triplet(s)}                 & \textbf{Statement}                                     & \textbf{API Write Call(s)}                                             \\
        \midrule
        $[\langle per_1, rel, org \rangle, \langle per_2, rel, org \rangle, ...]$      & [\texttt{per}$_1$, \texttt{per}$_2$, ...] is\textbackslash are \texttt{rel} to \texttt{org}.                        & \texttt{[MEM\_WRITE\{per$_1$\nextArg rel$_1$\nextArg org\}][MEM\_WRITE\{per$_2$\nextArg rel$_2$\nextArg org\}]...}      \\
        \bottomrule
    \end{tabular}}
\end{center}
\caption[]{Memory write data example structure for finetuning.}
\label{tab:mem_write_data_template}
\end{table*}

%% file: figures/eval_example_1.tex
\begin{figure*}[t]
\centering
    \includegraphics[width=0.93\linewidth, trim=0 0 0 0, clip] {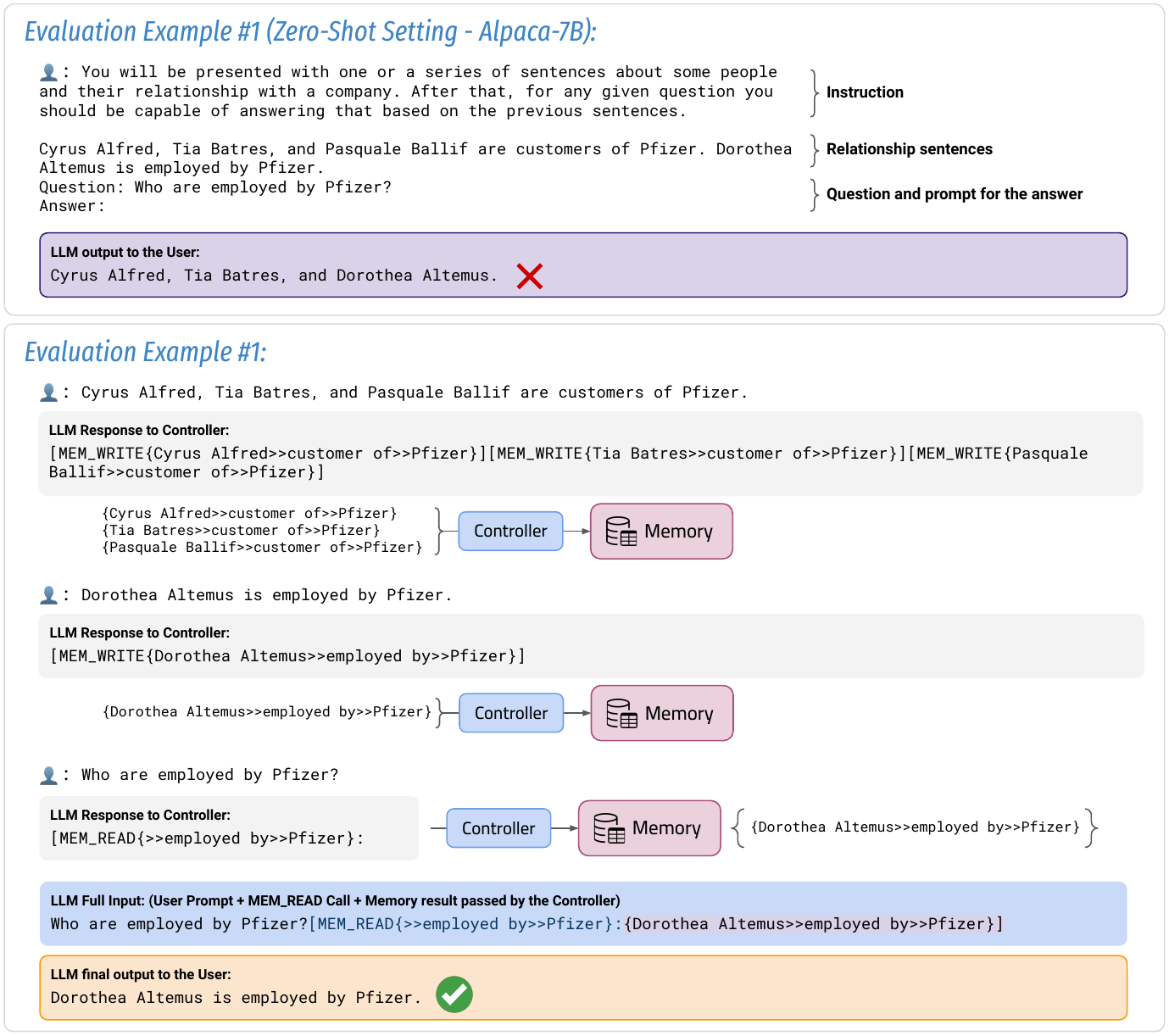}
    \caption{
        An example that has an incorrect result in a zero-shot setting and a correct one in our approach.
        Note that in the zero-shot setting the model has direct access to the information required for answering the question in its input and still end up with an incorrect answer.
        However, in our approach each of the user prompts could be given to the \framework{} in separately.
        Another example is mentioned in the appendix Figure \ref{fig:eval_example_2}.
    }
    \label{fig:eval_example_1}
\end{figure*}

%% file: figures/qa_example.tex
\begin{figure*}[h!]
\centering
    \includegraphics[width=0.93\linewidth, trim=0 0 0 0, clip] {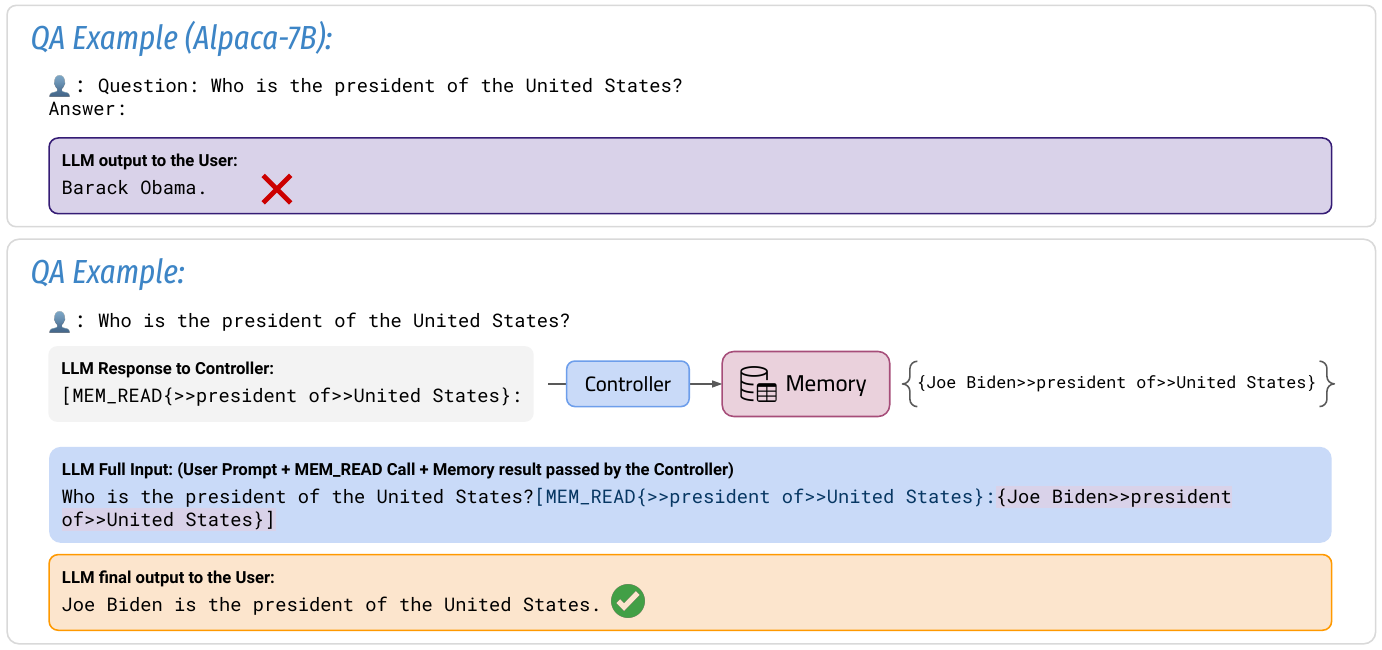}
    \caption{
        Asking a question which requires temporal context usually leads to an outdated answer as shown here with Alpaca.
        However, in our \framework{} with the aid of a modifiable memory, these questions could be answered by simply providing a updated memory entry.
    }
    \label{fig:qa_president_example}
\end{figure*}

%% file: figures/eval_example_2.tex
\begin{figure*}[t]
\centering
    \includegraphics[width=0.95\linewidth, trim=0 0 0 0, clip] {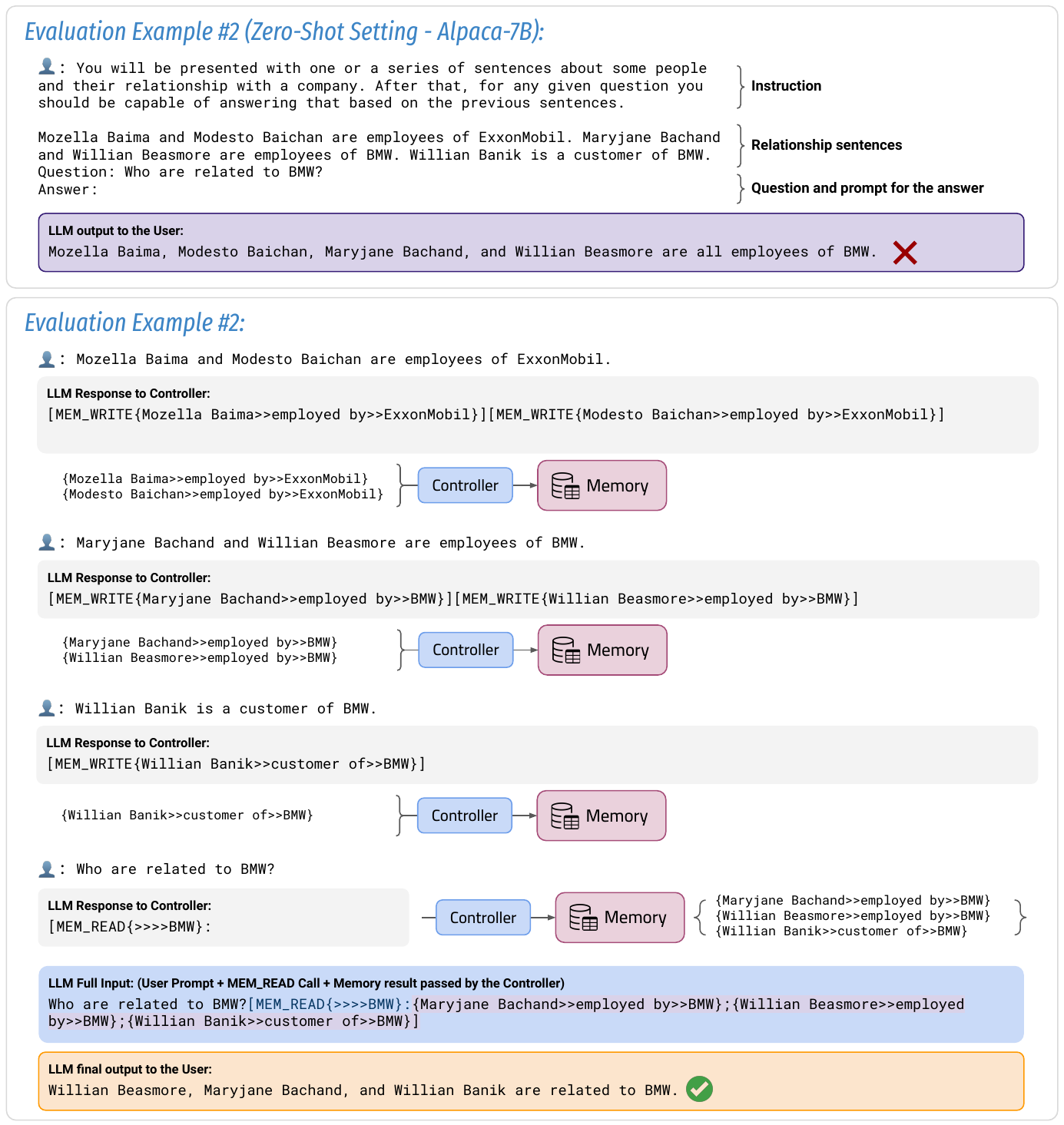}
    \caption{
        Another evaluation example that has an incorrect result in a zero-shot setting and a correct one in our approach.
    }
    \label{fig:eval_example_2}
\end{figure*}